\documentclass[runningheads]{llncs}
\usepackage{graphicx}
\usepackage{hyperref}
\usepackage{amsfonts,amssymb}
\usepackage{multirow}
\usepackage{booktabs}
\usepackage{bbding}
\begin{document}
	\title{Lesion-Aware Contrastive Representation Learning for Histopathology Whole Slide Images Analysis}
	\titlerunning{LACL for Histpathology WSI analysis}
	\author{Jun Li \inst{1} \and Yushan Zheng \inst{2(}\Envelope\inst{)} \and Kun Wu \inst{1} \and Jun Shi \inst{3(}\Envelope\inst{)} \and Fengying Xie \inst{1} \and \\ Zhiguo Jiang \inst{1}}
	% 1{Jun, Li} 
	% 2{Yushan, Zheng} 
	% 3{Kun, Wu}
	% 4{Jun, Shi} 
	% 5{Fengying, Xie} 
	% 6{Zhiguo, Jiang} 
	\authorrunning{J. Li et al.}
	\institute{Image Processing Center, School of Astronautics, Beihang University, Beijing, 102206, China. \and School of Engineering Medicine, Beijing Advanced Innovation Center on Biomedical Engineering, Beihang University, Beijing 100191, China.\\\email{yszheng@buaa.edu.cn} \and
	School of Software, Hefei University of Technology, Hefei 230601, China.}
	\maketitle              % typeset the header of the contribution
	\begin{abstract}
		Local representation learning has been a key challenge to promote the performance of the histopathological whole slide images analysis. The previous representation learning methods followed the supervised learning paradigm. However, manual annotation for large-scale WSIs is time-consuming and labor-intensive. Hence, the self-supervised contrastive learning has recently attracted intensive attention. The present contrastive learning methods treat each sample as a single class, which suffers from class collision problems, especially in the domain of histopathology image analysis. In this paper, we proposed a novel contrastive representation learning framework named Lesion-Aware Contrastive Learning (LACL) for histopathology whole slide image analysis. We built a lesion queue based on the memory bank structure to store the representations of different classes of WSIs, which allowed the contrastive model to selectively define the negative pairs during the training. Moreover, We designed a queue refinement strategy to purify the representations stored in the lesion queue. The experimental results demonstrate that LACL achieves the best performance in histopathology image representation learning on different datasets, and outperforms state-of-the-art methods under different WSI classification benchmarks. 
		The code is available at \url{https://github.com/junl21/lacl}.
		
	\end{abstract}
	\section{Introduction}
	In recent years, the research on the whole slide images (WSIs) analysis becomes popular in the field of digital pathology \cite{huang2021integration,lu2021data,vuong2021ranking}. The accuracy performance of the models for WSI analysis is crucial for the different downstream tasks, such as WSIs classification, retrieval, and survival prediction. However, due to the limitation of hardware resources, WSIs are difficult to directly fed into deep neural networks for training. The typical solution is to generate WSI-level representations by aggregating local tissue representations before inference \cite{huang2021integration,shao2021transmil}. In this situation, local representation learning is a key challenge to promote the development of digital pathology \cite{lu2021data,zheng2020diagnostic}. The previous methods proposed learning representations in the supervised learning paradigm based on pathologists' manual annotations \cite{arvaniti2018automated,han2018learning,jiang2019breast}, but the development of these methods has hit a bottleneck because of the expensive cost of large-scale annotations.
	
	To address this issue, transfer learning from ImageNet\cite{russakovsky2015imagenet} is a well-adopted  strategy to learn local representations\cite{lu2021data}. However,  The performance of these methods is limited by the semantic gap between the natural images and the histopathological images. Some weakly supervised learning methods tended to transfer the the feature domain from natural scene to histopathology by multi-instance learning (MIL), but the upper bound of such methods is limited by the performance of the representation model trained outside the histopathology domain \cite{ilse2018attention,lerousseau2020weakly,wang2019rmdl}. Moreover, MIL methods mainly focus on the positive vs. negative discrimination of the local representations, which makes it weak in describing the differences of subtype lesions.
	
	In this case, self-supervised contrastive learning has been introduced for histopathology image representation learning. It learns image patterns based on large-scale unlabeled data, which has demonstrated superiority to the ImageNet-trained model in a variety of downstream vision tasks \cite{chen2020simple,chen2021exploring,he2020momentum}. 
	There are several works applying contrastive learning methods into the field of computational pathology \cite{wang2021transpath,yang2021self}. 

	The success of the contrastive learning depends on the design of informative of the positive and negative pairs \cite{robinson2020contrastive}. However, the present sampling strategy introduces class collision problem\cite{zheng2021weakly} for the reason that it treats each instance as a single class. Specifically, common contrastive learning treats any two samples in the dataset as a negative pair, even though they belong to the same class semantically. It hurts the quality of the learned representation, especially for histopathological images where the image content is rather complex. Several works have tried to refine the sampling strategy\cite{tian2020makes,xiao2020should}, but there is lack of sampling strategies specifically designed for pathological images.
	
    In this paper, we rethought the applicability of contrastive learning in digital pathology, and proposed a weakly supervised representations learning method named lesion-aware contrastive learning (LACL).
    A dynamic queue was built to refine the negative pairs selection in the LACL. 
    Moreover, we designed a queue refinement strategy (QRS) to continuously purify queues during the training process. 
    The experimental results on two WSIs classification benchmarks show that the representations learned by our method achieves significant improvement in both accuracy and AUC metrics, when compared to the state-of-the-art representation learning methods.
    The contribution of this paper can be summarized into two aspects:
	\begin{itemize}
		\item[•] We propose a novel weakly supervised contrastive learning framework with a designed lesion queue. The lesion queue is a dynamic memory bank, which is used to replace the memory bank of MoCo \cite{he2020momentum} in storing the features from different types of WSIs. Based on the queue, the contrastive learning for each sample is restricted to the samples with different WSI labels, which enables the model to embed the WSI lesion information to the representations.
	\item[•] We design a queue refinement strategy (QRS). In each step of training, the typical samples for each class are selected by the QRS to update the queue, which is different from MoCo \cite{he2020momentum} that simply updates the memory bank using all the batch samples. It makes the representations stored in the queue more representative to the lesion types, and therefore improves the effectiveness for WSI analysis.
		\item[•] The experimental results show our LACL can outperform state-of-the-art representation learning methods for downstream whole slide images analysis.
	\end{itemize}
		
	\begin{figure}[t]
		\includegraphics[width=\textwidth]{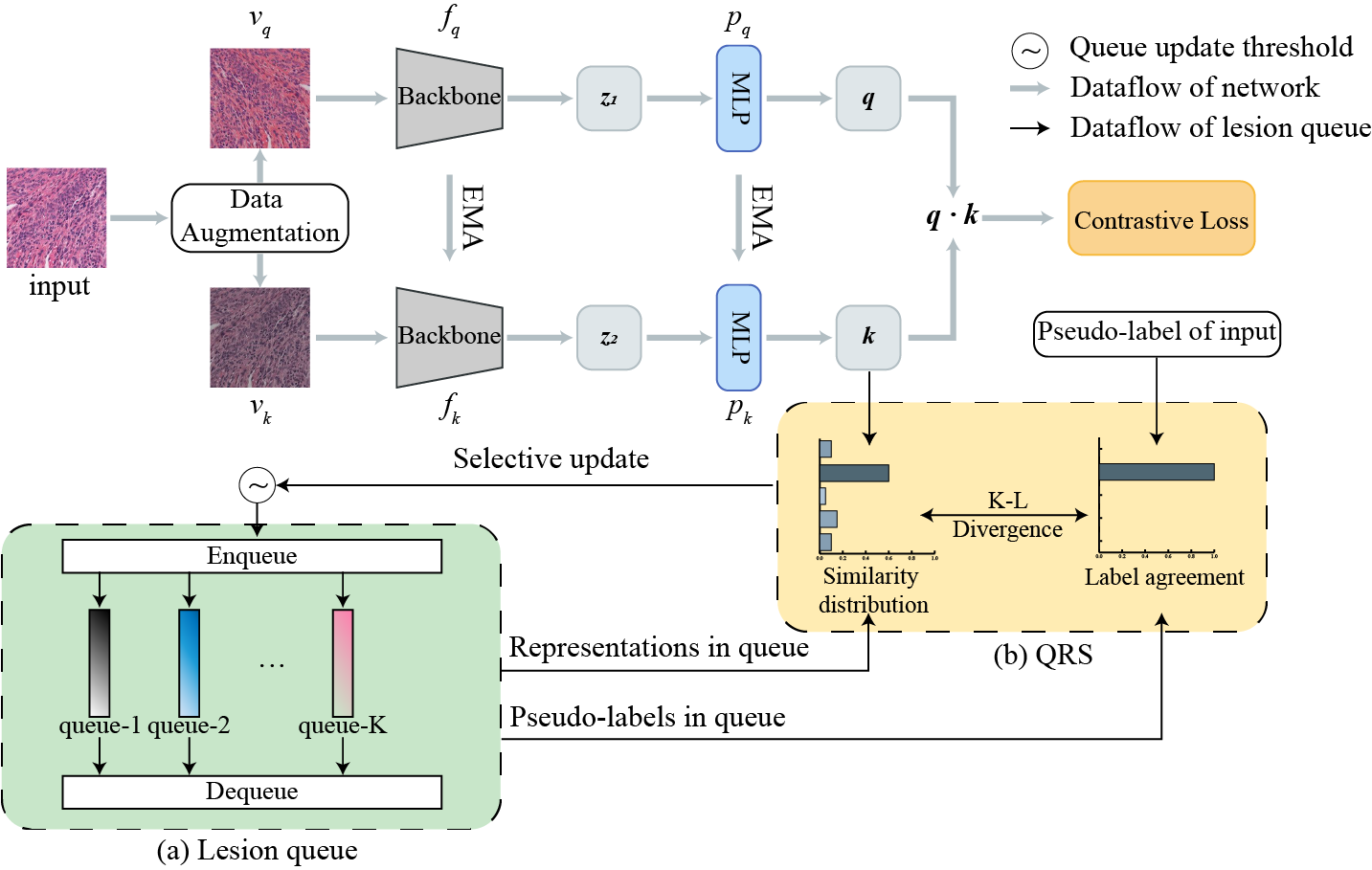}
		\caption{The overview of the proposed lesion-aware contrastive learning (LACL) framework which takes MoCo v2 as the basic structure. (a) is the proposed lesion queue redesigned from memory bank in MoCo. It allows the model to selective construct the contrastive sample pairs and enables the model to embed the WSI lesion information to the representations. (b) represents the queue refinement strategy (QRS), which is designed for selecting typical samples for each class to update the lesion queue.} \label{fig1}
	\end{figure}
	
	\section{Methods}
	The overview of the proposed framework is illustrated in Fig.~\ref{fig1}. The architecture of the framework is constructed based on MoCo v2, where we have redesigned the original memory bank and renamed it as lesion queue, as shown in Fig.~\ref{fig1}(a). In addition, a queue refinement strategy (QRS) is built for queue updating, as shown in Fig.~\ref{fig1}(b). The details of the proposed framework are described as followings.

	\subsection{Contrastive Learning baseline}
	Contrastive learning (CL) methods aim at mining intra-class similarities and inter-class differences from the content of images. The CL structure used in this paper is a siamese network consisting of two branches, namely a query branch and a key branch. The query branch consists of an encoder $f_q$ and a projector $p_q$, which are determined by a set of trainable parameters $\theta_q$. The key branch shares the same structure with the query branch that is represented as $f_k$ and  $p_k$ and determined by trainable parameters $\theta_k$. The difference is that the parameters of the key branch are updated by the exponential moving average (EMA) mechanism from the query branch by equation $\theta_k \gets m\theta_q+(1-m)\theta_k, m \in [0, 1)$. In this paper, ResNet50 is used as the encoder, and the projector is a multilayer perceptron (MLP) which is composed in order of a linear layer, a ReLU layer and another linear layer.
	
	Given an image $x$, two different augmented views for the image, $v_q$ and $v_k$, are fed into the query branch and the key branch, respectively. Firstly, $v_q$ and $v_k$ pass through the corresponding encoder to generate the representations $\textbf{z}_q = f_q(v_q)$ and $\textbf{z}_k = f_k(v_k)$. Then a projector $g(\cdot)$ is adopted to map the representations to the space where contrastive loss is applied. Here we defined the outputs of the projectors as $\textbf{q} = p_q(\textbf{z}_q)$ and $\textbf{k}_+ = p_k(\textbf{z}_k)$. Referring to MoCo, we define $\textbf{k}_+$ as the positive sample, and those representations from the memory bank as negative keys $\{\textbf{k}_0,\textbf{k}_1, \dots,\textbf{k}_N\}$, and the objective funtion is optimized by  minimizing the InfoNCE loss function
	\begin{equation}
	\label{eq1}
	\mathcal{L} = -\log\frac{\exp(\textbf{q}^{\mathrm{T}}\textbf{k}_+/\tau)}{\sum_{i=0}^{N} \exp(\textbf{q}^{\mathrm{T}}\textbf{k}_i/\tau)},
	\end{equation}
	where $\tau$ is a temperature hyper-parameter, and the sum is over one positive and $N$ negative samples.
	
	\subsection{Lesion Queue Construction}
	\label{section_QRS}
	It is notable that the contrastive representation learning in the majority works are achieved in unsupervised paradigm without any level of supervision from the WSI type information. It makes the representations hard to aware the semantic gaps of histopathology image content involving intra-class variation and inter-class similarity. This problem does harm to the downstream WSI analysis tasks.
	
	To tackle the problem, we propose to rebuilding the memory bank based on the WSI type information of the contrastive samples, and rename it as lesion queue. Specifically, we build a queue for each WSI class, and let each queue only stores the image representations obtained from its corresponding class of WSI. Based on the settings, we redesign the contrastive objective by dynamically select the negative samples for each positive sample with the modified InfoNCE loss function
	\begin{equation}
	\mathcal{L} = -\log\frac{\exp(\textbf{q}^{\mathrm{T}}\textbf{k}_+/\tau)}{\sum_{y\in \mathbb{C}}\sum_{i=0}^{M} \exp(\textbf{q}^{\mathrm{T}}\textbf{k}_{yi}/\tau)},
	\end{equation}
	where $M$ is the capacity of each queue, $\mathbb{C}$ is the set of classes that need to be contrasted as negative samples, and $\textbf{k}_{yi}$ denotes the $i$-th sample in the $y$-th queue.
	
	In this paper, we set a pseudo-label for a certain sample as the WSI label to which the sample belongs. Then, for a sample $x$ with pseudo-label $\tilde{y}$, we set $\mathbb{C} = \{y | y \neq \tilde{y}\}$. In this way, the type information of the samples are implicitly embedded into the contrastive learning process, which guides the image representations from the same lesion to allocate in the consistent direction of the feature space. It relives the class collision problems caused by the common contrastive representation learning and therefore will be more effective in representing histopathological images.

	\subsection{Queue Refinement Strategy}
	To ensure the samples in queues to be representative to the corresponding class during the training process, the queues need to be updated during the training. Here, we propose a novel queue refinement strategy (QRS). QRS purifies each queue with its most representative representations from each mini-batch by measuring the similarities between the to-be-updated representations and the queue representations.
	Specifically, we define $\mathcal{P}(x)$ to represent the similarity distribution between the to-be-updated $k_+$ and representations in queues, and $\mathcal{Q}(x)$ to represent the expected distribution with equations
	\begin{equation}
	\mathcal{P}(x_{yi}) = \frac{\exp(\textbf{k}_+^{\mathrm{T}}\textbf{k}_{yi})}{\sum_{y=1}^K\sum_{i=1}^M \exp(\textbf{k}_+^{\mathrm{T}}\textbf{k}_{yi})}
	\end{equation}
	
	\begin{equation}
	\mathcal{Q}(x_{yi}) = \frac{\exp(1(y=\tilde{y}))}{\sum_{y=1}^K\sum_{i=1}^M \exp(1(y=\tilde{y}))},
	\end{equation}
	where $K$ is total number of classes of the WSIs, $1(\cdot)$ takes $1$ when $y=\tilde{y}$, and $0$, otherwise.
	
	Then, we determine the collection of samples for updating basing on the consensus that the representation distance between the samples from the same class is smaller than those from the different classes. It can be achieved by examining the distribution consistency of $\mathcal{P}(x)$ and $\mathcal{Q}(x)$ with Kullback-Leibler (KL) divergence:
	
	\begin{equation}
	\mathbb{U} = \{\textbf{k}_+ | D_{KL}(\mathcal{P}(x)||\mathcal{Q}(x))\leq\frac{1}{|\mathbb{B}|}\sum_{x \in \mathbb{B}} D_{KL}(\mathcal{P}(x)||\mathcal{Q}(x))\},
	\end{equation}
	
	\begin{equation}
	D_{KL}(\mathcal{P}(x)||\mathcal{Q}(x)) = \sum_{y=1}^K\sum_{i=1}^M{\mathcal{P}(x_{yi})\log\frac{\mathcal{P}(x_{yi})}{\mathcal{Q}(x_{yi})}},
	\end{equation}
	where $\mathbb{U}$ is the set of representations which needs to be updated into the queue from each mini-batch, $\mathbb{B}$ is the mini-batch and $D_{KL}(\cdot)$ is the K-L divergence.

	\section{Datasets}
	To verify the proposed method, we collected two datasets including different tasks: EGFR mutation identification for lung cancer and endometrial subtyping, which are introduced below.
	
	\textbf{EGFR} is a dataset of lung adenocarcinoma whole slide images for epidermal growth factor receptor (EGFR) gene mutation identification. The dataset contains a total of 754 WSIs, which are categorized into 5 classes including EGFR 19del mutation, EGFR L858R mutation, none common driver mutations (wild type), other driver gene mutation, and cancer-free tissue (Normal).
	
	\textbf{Endometrium} Endometrium includes 698 histopathological WSIs of endometrial cases, which includes 5 categories, namely Normal, well differentiated endometrioid adenocarcinoma (WDEA), moderately differentiated endometrioid adenocarcinoma (MDEA), lowly differentiated endometrioid adenocarcinoma (LDEA), and Serous endometrial intraepithelial carcinoma (SEIC). 
		
	In each dataset, the WSIs were randomly split into train, validation and test parts following the ratio of 6:1:3 at the patient-level.
	The train part is used to train the models, and validation part is used to do early stop for the WSI classification benchmarks, and the final result is obtained within the test part. The WSIs were divided into non-overlapping patches in size of 256 $\times$ 256 for the representation learning and feature extraction. We assign a pseudo-label for each patch to the WSI label that the patch belongs.
	
	\section{Experiments and Results}
	\subsection{Experimental Setup}
	For each model involved in the experiment, we trained it by 100 epochs on the training set with the batch size as 256. We set the capacity of each lesion queue $M = 65536 / K$, to make the total length of the lesion queue is 65536 that is the same as MoCo's memory bank. Afterwards, we took over the backbone of the Resnet50 to extract features for the downstream WSI classification tasks. 
	
	In this paper, we evaluated the representation learning performance under two WSI classification benchmarks, CLAM\cite{lu2021data} and TransMIL\cite{shao2021transmil}.
	Both methods use the attention mechanism to aggregate features. The difference is that CLAM focuses on the parts of features in WSI that are most useful for diagnosis. While, TransMIL considers the spatial connections of the patches within the WSI and meanwhile extracts the long-term dependencies through the Transformer\cite{vaswani2017attention} architecture. For both of the datasets, we utilize evaluation metrics of accuracy, macro-average area under the curve (AUC) and macro-average F1-Score for the multi-class classification tasks.
	
	All the models were implemented in Python 3.8 with PyTorch 1.7.1 and run on a computer with 4 Nvidia GTX 2080Ti GPUs.
	
	\subsection{Structural Verification}
	
	\begin{table}[t]
		\centering
		\caption{Ablation study on the EGFR dataset and the Endometrium dataset under the TransMIL benchmark, where the accuracy (ACC), the macro-average area under the curve (AUC) and the macro-average F1-Score are used as metrics.}
		\label{tab1}
		\setlength{\tabcolsep}{5pt}
		\begin{tabular}{l|llc|llc}
			\toprule
			\multicolumn{1}{l|}{\multirow{2}{*}{\textbf{Methods}}} & \multicolumn{3}{c|}{\textbf{EGFR}} & \multicolumn{3}{c}{\textbf{Endometrium}}\\ 
			\multicolumn{1}{c|}{} & \multicolumn{1}{c}{ACC} & \multicolumn{1}{c}{AUC} & F1-Score & \multicolumn{1}{c}{ACC} & \multicolumn{1}{c}{AUC} & F1-Score \\ 
			\midrule
			LACL-w/o-queue (MoCov2)  & \multicolumn{1}{l}{0.534} & \multicolumn{1}{l}{0.826} & 0.517 & \multicolumn{1}{l}{0.436} & \multicolumn{1}{l}{0.670} & 0.259\\ 
			LACL-w/o-QRS   & \multicolumn{1}{l}{0.556} & \multicolumn{1}{l}{0.844} & 0.547 & \multicolumn{1}{l}{0.474} & \multicolumn{1}{l}{0.715} & 0.382\\ 
			%LACL-simplest & \multicolumn{1}{l|}{-} & - & \multicolumn{1}{l|}{-} & - \\ \hline
			LACL     & \multicolumn{1}{l}{0.574} & \multicolumn{1}{l}{0.855} & 0.552 & \multicolumn{1}{l}{0.488} & \multicolumn{1}{l}{0.732} & 0.390\\ 
			\bottomrule
		\end{tabular}
	\end{table}
	
	We set up ablation experiments to investigate the impact of our proposed method. The results are shown in the Table~\ref{tab1}.
	LACL-w/o-queue represents the contrastive learning without the proposed lesion queue and the corresponding QRS strategy. In this case, it degrades to the MoCov2. In LACL-w/o-QRS, we did not use QRS to update the lesion queue. Instead, we updated the queue with the strategy in MoCov2.
    Table~\ref{tab1} shows LACL-w/o-queue suffers from a decrease by 0.018 and 0.045 in AUC in the task for EGFR and Endometrium, respectively, compared to LACL-w/o-QRS. It demonstrates the effectiveness of the proposed lesion queue.
	When comparing the results of LACL-w/o-QRS and LACL, we found that the QRS  further improves the AUC by 0.011 and 0.017. It demonstrates the necessity for a QRS strategy.
	
	\subsection{Comparisons with State-of-the-Art Methods}
	We compared the representations under 6 different WSI presenting strategies. (1) ImageNet denotes extracting the patch representations using the ResNet50 pre-trained on the ImageNet dataset \cite{russakovsky2015imagenet}. (2) MoCov2 \cite{chen2020improved} is the widely applied contrastive learning framework, which is the basis of the proposed method. (3) SimSiam \cite{chen2021exploring} is another popular self-supervised learning method based on positive pair contrasting. (4) SimTriplet \cite{liu2021simtriplet} redesign the sampling strategy for positive pairs for pathological images, which allows the model to mine the relevance of adjacent tissues. (5) Pseudo-labels denote extracting the patch representations using the ResNet50 trained with pseudo-labels (WSI-level labels). (6) Lerousseau. et al. \cite{lerousseau2020weakly} is a weakly supervised patch representation learning and classification method.
	
    The results are shown in Tables~\ref{tab2} and \ref{tab3}.
	It shows that the negative-independent methods, i.e., SimSiam and SimTriplet, perform worse on the Endometrium than on the EGFR. The main reason is that the class imbalance of Endometrium dataset is more serious than the EGFR. It indicates that negative-independent methods suffer from the long-tailed distribution dataset. In contrast, the negative-dependent methods, including  MoCo and LACL, delivered better performance in accuracy and robustness. It is mainly because the negative pairs can provide more information to assist models in mining intra-class variation. At the same time, large-capacity memory bank could relieve the problem of class imbalance on models. It is obvious that the overall performance of SimTriplet on the EGFR dataset is significantly better than that on the Endometrium dataset. It may because the fact that the tissues of different differentiation stages usually intersect with each other, and thereby the usage of the adjacent tissues as positive pairs would introduce semantic ambiguity and hurt the quality of the representations. In comparison, the proposed LACL achieves consistent performance on the two datasets. It benefits from the design of equal-length lesion queue, and corresponding sampling strategy from the perspective of negative pairs.
	
	Overall, the proposed LACL achieves the best representation learning performance on both the EGFR dataset and Endometrium dataset under both the CLAM and the TransMIL benchmarks. It demonstrates that the design of the lesion queue and the corresponding QRS have provided more discriminative and robust image representations for WSI analysis.
    
	\begin{table}[t]
		\centering
		\caption{Comparisons with state-of-the-art methods on the EGFR dataset, where the metrics are the same as Table~\ref{tab1}}\label{tab2}
		\setlength{\tabcolsep}{5pt}
		\begin{tabular}{l|llc|llc}
			\toprule
			\multirow{2}{*}{\textbf{Methods}} &
			\multicolumn{3}{c|}{\textbf{CLAM}} &
			\multicolumn{3}{c}{\textbf{TransMIL}} \\ 
			&
			\multicolumn{1}{c}{ACC} &
			\multicolumn{1}{c}{AUC} &
			\multicolumn{1}{c|}{F1-Score} &
			\multicolumn{1}{c}{ACC} &
			\multicolumn{1}{c}{AUC} &
			F1-Score \\ 
			\midrule
			ImageNet\cite{russakovsky2015imagenet} &
			\multicolumn{1}{l}{0.435} & 
			\multicolumn{1}{l}{0.754} &
			\multicolumn{1}{c|}{0.429} &
			\multicolumn{1}{l}{0.502} &
			\multicolumn{1}{l}{0.798} &
			0.484 \\ 
			MoCo v2\cite{chen2020improved} &
			\multicolumn{1}{l}{0.462} &
			\multicolumn{1}{l}{0.803} &
			\multicolumn{1}{c|}{0.447} &
			\multicolumn{1}{l}{0.534} &
			\multicolumn{1}{l}{0.826} &
			0.517 \\
			SimSiam\cite{chen2021exploring} &
			\multicolumn{1}{l}{0.453} &
			\multicolumn{1}{l}{0.768} &
			\multicolumn{1}{c|}{0.391} &
			\multicolumn{1}{l}{0.502} &
			\multicolumn{1}{l}{0.801} &
			0.483 \\ 
			SimTriplet\cite{liu2021simtriplet} &
			\multicolumn{1}{l}{0.511} &
			\multicolumn{1}{l}{0.819} &
			\multicolumn{1}{c|}{0.464} &
			\multicolumn{1}{l}{0.529} &
			\multicolumn{1}{l}{0.841} &
			0.510 \\ 
			\midrule
			Pseudo-labels\ &
			\multicolumn{1}{l}{0.426} &
			\multicolumn{1}{l}{0.776} &
			\multicolumn{1}{c|}{0.403} &
			\multicolumn{1}{l}{0.466} &
			\multicolumn{1}{l}{0.802} &
			0.421 \\ 
			Lerousseau. et al.\cite{lerousseau2020weakly}\ &
			\multicolumn{1}{l}{0.390} &
			\multicolumn{1}{l}{0.710} &
			\multicolumn{1}{c|}{0.339} &
			\multicolumn{1}{l}{0.368} &
			\multicolumn{1}{l}{0.660} &
			0.326 \\ 
			\midrule
			LACL(Ours) &
			\multicolumn{1}{l}{\textbf{0.525}} &
			\multicolumn{1}{l}{\textbf{0.826}} &
			\multicolumn{1}{c|}{\textbf{0.510}} &
			\multicolumn{1}{l}{\textbf{0.574}} &
			\multicolumn{1}{l}{\textbf{0.855}} &
			\textbf{0.552} \\ 
			\bottomrule
		\end{tabular}
	\end{table}
	\begin{table}[t]
		\centering
		\caption{Comparisons with state-of-the-art methods on the Endometrium dataset, where the metrics are the same as Table~\ref{tab1}.}\label{tab3}
		\setlength{\tabcolsep}{5pt}
		\begin{tabular}{l|llc|llc}
			\toprule
			\multirow{2}{*}{\textbf{Methods}} &
			\multicolumn{3}{c|}{\textbf{CLAM}} &
			\multicolumn{3}{c}{\textbf{TransMIL}} \\ 
			&
			\multicolumn{1}{c}{ACC} &
			\multicolumn{1}{c}{AUC} &
			\multicolumn{1}{c|}{F1-Score} &
			\multicolumn{1}{c}{ACC} &
			\multicolumn{1}{c}{AUC} &
			F1-Score \\ 
			\midrule
			ImageNet\cite{russakovsky2015imagenet} &
			\multicolumn{1}{l}{0.483} &
			\multicolumn{1}{l}{0.675} &
			\multicolumn{1}{c|}{0.404} &
			\multicolumn{1}{l}{0.427} &
			\multicolumn{1}{l}{0.662} &
			0.258 \\ 
			MoCo v2\cite{chen2020improved} &
			\multicolumn{1}{l}{0.502} &
			\multicolumn{1}{l}{0.710} &
			\multicolumn{1}{c|}{0.369} &
			\multicolumn{1}{l}{0.436} &
			\multicolumn{1}{l}{0.670} &
			0.289 \\
			SimSiam\cite{chen2021exploring} &
			\multicolumn{1}{l}{0.483} &
			\multicolumn{1}{l}{0.663} &
			\multicolumn{1}{c|}{0.368} &
			\multicolumn{1}{l}{0.436} &
			\multicolumn{1}{l}{0.639} &
			0.233 \\ 
			SimTriplet\cite{liu2021simtriplet} &
			\multicolumn{1}{l}{0.422} &
			\multicolumn{1}{l}{0.639} &
			\multicolumn{1}{c|}{0.258} &
			\multicolumn{1}{l}{0.408} &
			\multicolumn{1}{l}{0.636} &
			0.181 \\ 
			\midrule
			Pseudo-labels\ &
			\multicolumn{1}{l}{0.417} &
			\multicolumn{1}{l}{0.670} &
			\multicolumn{1}{c|}{0.332} &
			\multicolumn{1}{l}{0.412} &
			\multicolumn{1}{l}{0.664} &
			0.326 \\ 
			Lerousseau. et al.\cite{lerousseau2020weakly}\ &
			\multicolumn{1}{l}{0.408} &
			\multicolumn{1}{l}{0.586} &
			\multicolumn{1}{c|}{0.196} &
			\multicolumn{1}{l}{0.427} &
			\multicolumn{1}{l}{0.579} &
			0.185 \\ 
			\midrule
			LACL(Ours) &
			\multicolumn{1}{l}{\textbf{0.531}} &
			\multicolumn{1}{l}{\textbf{0.756}} &
			\multicolumn{1}{c|}{\textbf{0.447}} &
			\multicolumn{1}{l}{\textbf{0.488}} &
			\multicolumn{1}{l}{\textbf{0.732}} &
			\textbf{0.390} \\ 
			\bottomrule
		\end{tabular}
	\end{table}

	\section{Conclusion}
	In this paper, we proposed a novel lesion-aware contrastive learning framework to generate robust and discriminative representations for the histopathological image analysis. 
	We built the lesion queue module to address the class collision problems in the common contrastive learning methods. The lesion queue enables the model to dynamically select negative pairs during the training process. 
	In addition, we designed a queue refinement strategy to refine the queue update process, which further improved the model performance.
	Experimental results on WSI sub-typing tasks have demonstrated the effectiveness and robustness of our proposed representation learning method.
	In future work, we will try to leverage the positive pair mining strategy to the proposed method, to further improve its performance for WSI analysis.
	
	\subsubsection*{Acknowledgments.}
	This work was partly supported by the National Natural Science Foundation of China [grant no. 61901018, 62171007, 61906058, and 61771031], partly supported by the Anhui Provincial Natural Science Foundation [grant no. 1908085MF210], and partly supported by the Fundamental Research Funds for the Central Universities of China [grant no. JZ2022HGTB0285].
	
	\bibliographystyle{splncs04}
	\bibliography{ref}

\end{document}